\documentclass[UTF8,fontset=none,a4paper,10pt]{ctexart}

\usepackage[a4paper,left=19mm,right=19mm,top=18mm,bottom=19mm,headheight=14pt]{geometry}
\usepackage{amsmath,amssymb,bm,mathtools}
\usepackage{booktabs,tabularx,array,longtable}
\usepackage{xcolor}
\usepackage{enumitem}
\usepackage{listings}
\usepackage{titlesec}
\usepackage{fancyhdr}
\usepackage{microtype}
\usepackage{hyperref}
\usepackage{bookmark}
\renewcommand{\abstractname}{Abstract}
\renewcommand{\contentsname}{Contents}
\renewcommand{\figurename}{Figure}
\renewcommand{\tablename}{Table}
\renewcommand{\refname}{References}
\renewcommand{\appendixname}{Appendix}

\definecolor{navy}{HTML}{17365D}
\definecolor{blue}{HTML}{2459A6}
\definecolor{lightblue}{HTML}{EAF1FB}
\definecolor{linegray}{HTML}{D7DFEA}
\definecolor{codebg}{HTML}{F4F6F8}
\definecolor{codegreen}{HTML}{26734D}

\hypersetup{
  colorlinks=true,
  linkcolor=blue,
  citecolor=blue,
  urlcolor=blue,
  bookmarksnumbered=true,
  pdfauthor={Jiguo Li},
pdftitle={Position Encoding in Transformers: From Absolute and Relative Methods to Rotary Position Embeddings and Long-Context Scaling},
pdfsubject={Position encoding, rotary position embeddings, and long-context scaling}
}

\titleformat{\section}{\Large\bfseries\color{navy}}{\thesection}{0.7em}{}
\titleformat{\subsection}{\large\bfseries\color{navy}}{\thesubsection}{0.65em}{}
\titleformat{\subsubsection}{\normalsize\bfseries\color{navy}}{\thesubsubsection}{0.55em}{}
\titlespacing*{\section}{0pt}{14pt}{5pt}
\titlespacing*{\subsection}{0pt}{10pt}{3pt}
\titlespacing*{\subsubsection}{0pt}{7pt}{2pt}

\setlist[itemize]{leftmargin=1.8em,itemsep=1pt,topsep=2pt,parsep=0pt}
\setlist[enumerate]{leftmargin=2em,itemsep=1pt,topsep=2pt,parsep=0pt}

\newcommand{\term}[1]{\textbf{#1}}
\newcommand{\R}{\mathbb{R}}
\newcommand{\softmax}{\operatorname{softmax}}
\newcommand{\Attn}{\operatorname{Attn}}
\newcommand{\noteBox}[1]{%
  \begin{center}
  \fcolorbox{blue}{lightblue}{\parbox{0.93\linewidth}{#1}}
  \end{center}}
\setCJKsansfont[AutoFakeBold=2]{gbsn00lp.ttf}
\setCJKmonofont{gbsn00lp.ttf}
\makeatletter
\let\BilingualOriginalAddContentsLine\addcontentsline
\def\BilingualTocExtension{entoc}
\renewcommand{\addcontentsline}[3]{%
  \def\BilingualRequestedExtension{#1}%
  \def\BilingualStandardToc{toc}%
  \ifx\BilingualRequestedExtension\BilingualStandardToc
    \BilingualOriginalAddContentsLine{\BilingualTocExtension}{#2}{#3}%
  \else
    \BilingualOriginalAddContentsLine{#1}{#2}{#3}%
  \fi
}
\renewcommand{\tableofcontents}{%
  \section*{\contentsname
    \@mkboth{\MakeUppercase\contentsname}{\MakeUppercase\contentsname}}%
  \@starttoc{\BilingualTocExtension}%
}
\newcommand{\BilingualUseEnglishToc}{\gdef\BilingualTocExtension{entoc}}
\newcommand{\BilingualUseChineseToc}{\gdef\BilingualTocExtension{zhtoc}}
\def\BilingualAnchorPrefix{en}
\let\BilingualOriginalAppendix\appendix
\renewcommand{\appendix}{%
  \BilingualOriginalAppendix
  \renewcommand*{\theHsection}{\BilingualAnchorPrefix.appendix.\Alph{section}}%
}
\makeatother

\begin{document}
\renewcommand{\abstractname}{Abstract}
\renewcommand{\contentsname}{Contents}
\renewcommand{\figurename}{Figure}
\renewcommand{\tablename}{Table}
\renewcommand{\refname}{References}
\renewcommand{\appendixname}{Appendix}
\BilingualUseEnglishToc
\renewcommand*{\theHsection}{en.\arabic{section}}
\renewcommand*{\theHsubsection}{en.\arabic{section}.\arabic{subsection}}
\renewcommand*{\theHsubsubsection}{en.\arabic{section}.\arabic{subsection}.\arabic{subsubsection}}
\renewcommand*{\theHequation}{en.\arabic{equation}}
\renewcommand*{\theHfigure}{en.\arabic{figure}}
\renewcommand*{\theHtable}{en.\arabic{table}}
\gdef\BilingualAnchorPrefix{en}
\markboth{}{}
\pdfbookmark[0]{English Version}{bilingual.english}
\begin{center}
{\zihao{1}\bfseries\color{navy} Position Encoding in Transformers}\par
  \vspace{4pt}
{\zihao{3}\color{navy} From Absolute and Relative Methods to RoPE and Long-Context Scaling}\par
  \vspace{10pt}
{\normalsize\textbf{Jiguo Li\footnote{This report was completed with the assistance of Codex.}}}\par
  \vspace{2pt}
  {\small\href{mailto:jiguolee@gmail.com}{jiguolee@gmail.com}}
\end{center}
\vspace{5pt}
\hrule

\begin{abstract}
Self-attention models content-dependent interactions between tokens but does not by itself encode token order. Position encoding addresses this limitation by introducing absolute coordinates, relative distances, or position-dependent rotations into Transformer representations and attention scores. This technical survey develops a unified account of sinusoidal and learned absolute position embeddings, Shaw-style relative position representations, Transformer-XL, T5 relative position bias, ALiBi, and Rotary Position Embeddings (RoPE). We derive how RoPE converts absolute position indices into relative phase differences in Query-Key inner products and compare these methods in terms of where position is injected, computational cost, compatibility with KV caching, and length extrapolation. We then examine long-context extensions, including Position Interpolation, RoPE scaling laws, NTK-aware scaling, Dynamic NTK, NTK-by-parts, YaRN, LongRoPE, and LongRoPE2, with emphasis on frequency allocation, attention rescaling, training length, and target context length. We also summarize implementation considerations, evaluation protocols, and position-encoding choices in representative large language models. A central conclusion is that the ability to compute positional features beyond the training length does not imply reliable long-context generalization; context extension must be evaluated through short-context retention, position-wise perplexity, retrieval, reasoning, and long-context code tasks.
\end{abstract}


\noteBox{\textbf{Key point:} Absolute methods assign coordinates to tokens, relative methods represent pairwise distances, and RoPE writes position into the rotational phase of Query and Key vectors.}

\clearpage
\begingroup
\small
\setlength{\parskip}{0pt}
\linespread{1.02}\selectfont
\setcounter{tocdepth}{2}
\tableofcontents
\endgroup
\clearpage

\section{Why Transformers Need Explicit Positional Information}

\subsection{RNNs, CNNs, and Transformers}

RNN recursively in time:
\begin{equation}
  h_t=f(x_t,h_{t-1}),
  \label{en:eq:rnn}
\end{equation}
The calculation path itself is $x_1\rightarrow x_2\rightarrow\cdots\rightarrow x_n$, and the order naturally exists in state transfer. The convolution kernel of CNN first aggregates local neighborhoods and therefore also has local structure priors.

Transformer removes recursion and convolution, allowing all tokens to interact in parallel. This is the source of high throughput, and also means that the model structure itself no longer knows the order of tokens. The original Transformer must therefore additionally inject position information \cite{en:vaswani2017}.

\subsection{What does Self-Attention without position see?}

Given an input matrix:
\begin{equation}
  X=[x_1,x_2,\ldots,x_n]^\top\in\R^{n\times d},
  \label{en:eq:input}
\end{equation}
Standard Self-Attention calculation:
\begin{align}
  Q&=XW_Q,\quad K=XW_K,\quad V=XW_V, \label{en:eq:qkv}\\
  \Attn(X)&=\softmax\!\left(\frac{QK^\top}{\sqrt{d_h}}\right)V. \label{en:eq:attention}
\end{align}
The score of the $i$ Query for the $j$ Key is:
\begin{equation}
  s_{ij}=\frac{q_i^\top k_j}{\sqrt{d_h}}.
  \label{en:eq:score}
\end{equation}
The score uses the content vector but not the location $i,j$ directly. Assume $P$ is any permutation matrix, then:
\begin{equation}
  \Attn(PX)=P\Attn(X).
  \label{en:eq:permutation}
\end{equation}
Therefore, Self-Attention without the position mechanism is \term{ replacing the equivalent variable }: after the input is scrambled, the output is only scrambled in the same way. Position can be injected in the input representation, attention score or geometric transformation of Query/Key, corresponding to three types of solutions: absolute position, relative position and RoPE respectively.

\section{Overview of the Technical Evolution}

\begin{table}[htbp]
\centering
\small
\begin{tabularx}{\linewidth}{@{}p{1.2cm}p{3.5cm}X@{}}
\toprule
Time & represents work & core position mechanism \\
\midrule
2017 & Transformer & Fixed sine -- Cosine absolute position encoding \\
2018 & BERT & can learn absolute position embedding \\
2018 & Shaw et al. & Add relative distance representation to attention \\
2019 & Transformer-XL & Relative position decomposition, support fragment-level memory \\
2020 & T5 & Each attention head learns bucket distance bias \\
2021 & RoFormer / RoPE & Rotate by position Query and Key \\
2021 & ALiBi & Add linear distance penalty to logits \\
2023 & PI / YaRN & RoPE interpolation, frequency division scaling and attention calibration \\
2024--2025 & LongRoPE / LongRoPE2 & Search for dimension-wise scaling factors and mix long and short context training \\
\bottomrule
\end{tabularx}
\caption{Representative evolution of the Transformer position mechanism.}
\label{en:tab:timeline}
\end{table}

These methods are not a simple replacement of old with new. Fixed-length encoder, encoder-decoder and autoregressive LLM have different task structures, reasoning methods and contextual requirements, so multiple solutions coexist to this day.

\section{First generation: absolute position encoding}

Absolute position encoding provides the vector $p_i$ for position $i$:
\begin{equation}
  h_i^{(0)}=e(x_i)+p_i.
  \label{en:eq:absolute}
\end{equation}
It is equivalent to labeling each token with a global coordinate such as "position 137".

\subsection{Learned Absolute Position Embeddings}

The most straightforward implementation is to maintain the parameter table:
\begin{equation}
  P\in\R^{L_{\max}\times d}.
  \label{en:eq:learned-table}
\end{equation}
Line $i$ is the vector $p_i$ at position $i$. BERT adds token, segment and position embedding as input \cite{en:devlin2019}. It is simple to implement and can directly fit the position pattern within the task, but the parameter table is bound to a maximum length; untrained positions have no reliable representation; adjacent position differences are not structurally constrained; content and position are mixed before the first layer:
\begin{equation}
  q_i=(e(x_i)+p_i)W_Q=e(x_i)W_Q+p_iW_Q.
  \label{en:eq:absolute-proj}
\end{equation}

\subsection{Sine-cosine position encoding}

The original Transformer uses multiple sets of sine and cosine functions to generate the position vector \cite{en:vaswani2017}:
\begin{align}
  PE(pos,2i)&=\sin\!\left(\frac{pos}{10000^{2i/d_{\text{model}}}}\right), \label{en:eq:sin-pe}\\
  PE(pos,2i+1)&=\cos\!\left(\frac{pos}{10000^{2i/d_{\text{model}}}}\right). \label{en:eq:cos-pe}
\end{align}
Define the $i$ group angular frequency:
\begin{equation}
  \theta_i=10000^{-2i/d_{\text{model}}},
  \label{en:eq:sin-theta}
\end{equation}
Then a set of two-dimensional representation is:
\begin{equation}
  PE_i(pos)=
  \begin{bmatrix}
    \sin(pos\theta_i)\\
    \cos(pos\theta_i)
  \end{bmatrix}.
  \label{en:eq:sin-pair}
\end{equation}
The complete position vector concatenates multiple frequency pairs:
\begin{equation}
  PE(pos)=[\sin(pos\theta_0),\cos(pos\theta_0),\ldots,
  \sin(pos\theta_{d/2-1}),\cos(pos\theta_{d/2-1})].
  \label{en:eq:sin-full}
\end{equation}

\subsubsection{An intuitive explanation of "multiple clocks"}

Each pair of dimensions can be viewed as a pointer on the unit circle, and its phase and wavelength are:
\begin{align}
  \phi_i(pos)&=pos\theta_i, \label{en:eq:phase}\\
  \lambda_i&=\frac{2\pi}{\theta_i}=2\pi\cdot10000^{2i/d_{\text{model}}}. \label{en:eq:wavelength}
\end{align}
The large $\theta_i$ corresponds to the "fast clock", which can distinguish adjacent positions; the small $\theta_i$ corresponds to the "slow clock", which describes longer scales. The cycle of a single clock is repeated, and the combination of multiple clocks with different rotation speeds can distinguish positions in a wider range. The constant $10000$ is the frequency base, not the only mathematically correct choice.

When $d_{\text{model}}=8$, the four groups of frequencies are:
\begin{equation}
  \theta_0=1,\qquad \theta_1=0.1,\qquad
  \theta_2=0.01,\qquad \theta_3=0.001.
  \label{en:eq:d8-frequency}
\end{equation}

\subsubsection{Why Use Both Sine and Cosine?}

From the sum angle formula:
\begin{equation}
  \sin((pos+\Delta)\theta)=
  \sin(pos\theta)\cos(\Delta\theta)+
  \cos(pos\theta)\sin(\Delta\theta).
  \label{en:eq:addition}
\end{equation}
Saving only sin cannot obtain the translation result through fixed linear transformation; after saving in pairs, there are:
\begin{equation}
  PE_i(pos+\Delta)=
  \begin{bmatrix}
    \cos(\Delta\theta_i)&\sin(\Delta\theta_i)\\
    -\sin(\Delta\theta_i)&\cos(\Delta\theta_i)
  \end{bmatrix}PE_i(pos).
  \label{en:eq:sin-shift}
\end{equation}
The matrix depends only on the displacement $\Delta$. The dot product of two position codes at the same frequency is:
\begin{equation}
  PE_i(m)^\top PE_i(n)=\cos((m-n)\theta_i),
  \label{en:eq:sin-dot}
\end{equation}
The complete encoding satisfies:
\begin{equation}
  PE(m)^\top PE(n)=\sum_i\cos((m-n)\theta_i).
  \label{en:eq:sin-full-dot}
\end{equation}
Therefore, the sinusoidal encoding is an absolute position representation in form, but the geometric relationship includes relative displacement. The original Transformer actually uses the formula ~\eqref{en:eq:absolute} and does not directly use the formula ~\eqref{en:eq:sin-full-dot} as the attention score. The model still needs to be trained to take advantage of this structure.

\noteBox{\textbf{ It should be noted that the } position function can be calculated to any $pos$, which does not mean that the entire model can reliably understand any long context. The training length still constrains attention distribution, state representation and optimization results. }

\section{Second generation: relative position encoding}

Many relationships in the language and code are more concerned with $j-i$ than knowing $i$ versus $j$ respectively. For example, "the closest definition on the left", "the declaration closest to the call point", "the parameters of the two tokens on the right".

\subsection{Shaw: Add relative distance to Key and Value}

Shaw et al. directly added the relative distance of token pairs to Self-Attention\cite{en:shaw2018}. Cut the distance first:
\begin{equation}
  r_{ij}=\operatorname{clip}(j-i,-K,K),
  \label{en:eq:clip}
\end{equation}
Learn the relative Key vector $a^K_{r_{ij}}$ for each distance:
\begin{equation}
  s_{ij}=\frac{q_i^\top(k_j+a^K_{r_{ij}})}{\sqrt{d_h}}
  =\frac{q_i^\top k_j+q_i^\top a^K_{r_{ij}}}{\sqrt{d_h}}.
  \label{en:eq:shaw-score}
\end{equation}
The first item represents content matching, and the second item represents the current Query's preference for a certain direction and distance. Value can also be expressed in relative terms:
\begin{equation}
  z_i=\sum_j\alpha_{ij}(v_j+a^V_{r_{ij}}).
  \label{en:eq:shaw-value}
\end{equation}

\subsection{Transformer-XL: Relative positions in memory across segments}

Transformer-XL allows the current fragment to reuse the previous fragment's hidden state \cite{en:dai2019}. If the cache carries the absolute coordinates of the old fragment, reusing it to the new fragment will cause ambiguity; the relative position only depends on the true distance between the current Query and the historical Key. Its unnormalized attention can be decomposed into:
\begin{equation}
  \begin{aligned}
  A_{ij}={}&q_i^\top k_j+q_i^\top W_{k,R}R_{i-j}\\
           &+u^\top k_j+v^\top W_{k,R}R_{i-j},
  \end{aligned}
  \label{en:eq:txl}
\end{equation}
Corresponding to content -- content, content -- position, global content offset and global position offset respectively.

\subsection{T5: compress relative position into scalar bias}

T5 learns a scalar \cite{en:raffel2020} for each attention head and distance bucket:
\begin{equation}
  s_{ij}^{(h)}=
  \frac{q_i^{(h)\top}k_j^{(h)}}{\sqrt{d_h}}
  +b_{h,\operatorname{bucket}(j-i)}.
  \label{en:eq:t5-bias}
\end{equation}
Fine-grained bucketing is used when the distance is small, and logarithmic bucketing is approximated when the distance is large. Different heads can learn different distance priors, forming a clear decomposition of "content similarity $+$ distance preference".

The traditional relative position method is intuitive in modeling, but usually has to deal with $n\times n$ token pairs. Relative information may enter Key, Value, logits or multiple cross-terms, increasing the complexity of operator fusion, incremental decoding and efficient attention implementation. The key value of RoPE is to explicitly express relative displacements while retaining the standard $QK^\top$ form.

\section{Third Generation: Rotary Position Encoding RoPE}

RoPE does not add position vectors to the input, nor does it explicitly construct relative vectors for token pairs. Instead, it rotates Query and Key\cite{en:su2021} based on their absolute positions.

\subsection{Derivation of core formulas from two-dimensional rotations}

The two-dimensional rotation matrix is:
\begin{equation}
  R(\phi)=
  \begin{bmatrix}
    \cos\phi&-\sin\phi\\
    \sin\phi&\cos\phi
  \end{bmatrix}.
  \label{en:eq:rotation}
\end{equation}
The rotation angle at position $m$ is $m\theta$:
\begin{equation}
  q'_m=R(m\theta)q_m,\qquad k'_n=R(n\theta)k_n.
  \label{en:eq:rope-rotate}
\end{equation}
The rotated dot product is:
\begin{equation}
  \begin{aligned}
  q_m'^\top k_n'
  &=q_m^\top R(m\theta)^\top R(n\theta)k_n\\
  &=q_m^\top R((n-m)\theta)k_n.
  \end{aligned}
  \label{en:eq:rope-core}
\end{equation}
Used here:
\begin{equation}
  R(m\theta)^\top=R(-m\theta),\qquad
  R(-m\theta)R(n\theta)=R((n-m)\theta).
  \label{en:eq:rotation-group}
\end{equation}
The single vector $q'_m$ depends on the absolute position $m$; the Query-Key interaction only depends on the relative position through $n-m$. Therefore, \term{RoPE encodes the absolute position on a single vector, and } presents the relative position in the dot product.

\subsection{High-dimensional RoPE and complex-number forms}

Combine the head dimensions into two-dimensional subspaces, and the $r$ pair frequency is:
\begin{equation}
  \theta_r=b^{-2r/d_h},\qquad r=0,1,\ldots,d_h/2-1,
  \label{en:eq:rope-frequency}
\end{equation}
Classic base $b=10000$. The overall rotation matrix is:
\begin{equation}
  R_m=\operatorname{diag}\left(
  R(m\theta_0),R(m\theta_1),\ldots,R(m\theta_{d_h/2-1})
  \right).
  \label{en:eq:rope-block}
\end{equation}
If the two-dimensional vector is written as a complex number $z_r=x_{2r}+\mathrm{i}x_{2r+1}$, RoPE is equivalent to:
\begin{equation}
  z'_r=z_r e^{\mathrm{i}m\theta_r}.
  \label{en:eq:rope-complex}
\end{equation}
The phase factor multiplied by Query and Key is:
\begin{equation}
  e^{-\mathrm{i}m\theta_r}e^{\mathrm{i}n\theta_r}
  =e^{\mathrm{i}(n-m)\theta_r},
  \label{en:eq:rope-relative-phase}
\end{equation}
That is, only the relative phase remains after the absolute phase is subtracted.

\subsection{Engineering implementation and pairing conventions}

The following implementation uses adjacent dimension pairing. Some frameworks pair the first half dimension with the second half dimension; the two only have different arrangement conventions, but the weights, cos/sin cache, and rotation functions must be consistent.

\begin{lstlisting}
def apply_rope(x, cos, sin):
    x_even = x[..., 0::2]
    x_odd  = x[..., 1::2]
    y_even = x_even * cos - x_odd * sin
    y_odd  = x_even * sin + x_odd * cos

    y = torch.empty_like(x)
    y[..., 0::2] = y_even
    y[..., 1::2] = y_odd
    return y
\end{lstlisting}

Value is usually not rotated because position mainly affects the weight of "who to follow"; the aggregation object is still the content Value.

\subsection{RoPE and KV Cache}

Under fixed RoPE configuration, historical keys are usually cached in rotated form:
\begin{equation}
  k'_i=R_i k_i.
  \label{en:eq:cached-key}
\end{equation}
When generating location $t$ just calculate:
\begin{equation}
  q'_t=R_tq_t,\qquad k'_t=R_tk_t.
  \label{en:eq:decode-rope}
\end{equation}
Historical keys do not need to be recalculated. The most error-prone thing in engineering is position offset: prefill, decode, padding, sequence packing, prefix cache and sliding window must use the same logical position id for the same token.

\subsection{Does RoPE Guarantee Monotonic Attention Decay with Distance?}

Relative terms for a single frequency include:
\begin{equation}
  \cos((n-m)\theta_r),\qquad \sin((n-m)\theta_r),
  \label{en:eq:rope-oscillation}
\end{equation}
They oscillate periodically with distance and do not decay monotonically. RoFormer discussed the long-distance attenuation tendency \cite{en:su2021} after multi-frequency aggregation, but it cannot be deduced that "the further away the score must be, the smaller it will be" for any fixed Query/Key, any head and any distance.

\section{Extending RoPE to Long Contexts: From Uniform Scaling to Per-Dimension Search}

\subsection{Source of the problem: Extrapolation introduces untrained phases}

Let the original training length be $L_0$ and the target length be $L_1=sL_0$, where $s>1$. The period of the $r$th two-dimensional subspace is:
\begin{equation}
  T_r=\frac{2\pi}{\theta_r}=2\pi b^{2r/d_h}.
  \label{en:eq:rope-period}
\end{equation}
The high-frequency dimension goes through many cycles within $L_0$, and the local pattern is fully trained; the low-frequency dimension may not even cover a complete cycle. It is directly inferred that $L_1$ will cause some dimensions to enter an unseen phase interval, causing RoPE OOD. LongRoPE2 defines the theoretical critical dimension as the boundary \cite{en:longrope2} of whether the period exceeds the original training length:
\begin{equation}
  r_{\mathrm{crit}}=\min\{r:T_r\ge L_0\}.
  \label{en:eq:critical-dim}
\end{equation}
What the extension method really needs to deal with is the scaling method of each frequency and the adaptation of the model weights to the new phase distribution, not just increasing the value range of position id.

\subsection{RoPE scaling law: base, training length and extrapolation range}

Liu et al. studied RoPE length extrapolation from the perspective of period coverage and proposed RoPE-based extrapolation scaling laws\cite{en:liu2024scalingrope}. Here, ``scaling law'' describes the relationship among the rotation base, training length, and extrapolation range; it is distinct from classical model scaling laws relating parameter count, data volume, and compute.

For the original base $10000$, the period of the $n$ group frequency is:
\begin{equation}
  T_n=\frac{2\pi}{\theta_n}=2\pi\cdot10000^{2n/d_h}.
  \label{en:eq:scaling-period}
\end{equation}
If the pre-training length is $T_{\mathrm{train}}$, the paper writes the upper limit of the dimension that can cover at least a complete period within the training interval as:
\begin{equation}
  d_{\mathrm{extra}}
  =2\left\lceil
  \frac{d_h}{2}\log_{10000}\!\left(\frac{T_{\mathrm{train}}}{2\pi}\right)
  \right\rceil.
  \label{en:eq:critical-dimension-scaling-law}
\end{equation}
Low-frequency dimensions exceeding $d_{\mathrm{extra}}$ are not observed in full cycles during training and are more prone to phase distribution shifts beyond the training length. Taking $d_h=128$ and $T_{\mathrm{train}}=4096$ of LLaMA 2 as an example, the paper calculated $d_{\mathrm{extra}}=92$; the remaining 36 dimensions are regarded as the part \cite{en:liu2024scalingrope} that is more prone to instability during extrapolation.

When the rotation base is changed to $\beta>10000$ during the fine-tuning phase, this work uses the new period estimate of the critical dimension to extrapolate the upper limit:
\begin{equation}
  T_{\mathrm{extra}}
  =2\pi\,\beta^{d_{\mathrm{extra}}/d_h}.
  \label{en:eq:scaling-law-upper-bound}
\end{equation}
In turn, given the desired length $\widetilde T_{\mathrm{extra}}$, the required critical cardinality can be estimated by the relationship given in the paper:
\begin{equation}
  \beta_0
  =10000^{\log_{T_{\mathrm{train}}/(2\pi)}
  \left(\widetilde T_{\mathrm{extra}}/(2\pi)\right)}.
  \label{en:eq:scaling-law-critical-base}
\end{equation}

The paper also observes that, with a fixed fine-tuning length, turning the base smaller may also improve extrapolation, since more dimensions can experience fuller changes in the trigonometric function within the training interval. The corresponding three phase coverage points are:
\begin{equation}
  \beta_1=\frac{2T_{\mathrm{train}}}{\pi},\qquad
  \beta_2=\frac{T_{\mathrm{train}}}{\pi},\qquad
  \beta_3=\frac{T_{\mathrm{train}}}{2\pi}.
  \label{en:eq:scaling-law-small-base}
\end{equation}
This set of results reveals the connection between RoPE extrapolation and period coverage, but it is not appropriate to regard the formula ~\eqref{en:eq:scaling-law-upper-bound} as a hard contextual upper limit across models. The paper is mainly verified on LLaMA 2. The actual effective length is also affected by the long-distance dependency distribution in the corpus, fine-tuning data, attention head behavior and evaluation tasks.

\subsection{Position Interpolation: Uniformly Compressing Position Indices}

PI presses the position in the target range back to the original training interval \cite{en:chen2023pi}:
\begin{equation}
  m'=\frac{m}{s}=m\frac{L_0}{L_1}.
  \label{en:eq:pi-position}
\end{equation}
Equivalently, scale all frequencies uniformly:
\begin{equation}
  \theta'_r=\frac{\theta_r}{s}.
  \label{en:eq:pi-frequency}
\end{equation}
The advantage is that all new positions fall back into the trained phase range, with minimal structural changes; the disadvantage is that high frequencies are also compressed, the phase difference of adjacent tokens changes from $\theta_r$ to $\theta_r/s$, and the local resolution decreases. The expansion of the PI paper from 2K to 32K requires only short-range fine-tuning to adapt, but it is also clearly stated that there will be a certain performance loss within the original length \cite{en:chen2023pi}.

\subsection{NTK-aware: Non-uniform frequency scaling by modifying RoPE base}

The intuition of NTK-aware is: instead of compressing all frequencies equally like PI, adjust the frequency base $b$ so that high frequencies are maintained as much as possible and low frequencies bear more expansion. A common form sets the new base as:
\begin{equation}
  b'=b\,s^{d_h/(d_h-2)},
  \label{en:eq:ntk-base}
\end{equation}
thereby:
\begin{equation}
  \theta'_r=(b')^{-2r/d_h}
  =\theta_r\,s^{-2r/(d_h-2)}.
  \label{en:eq:ntk-frequency}
\end{equation}
The highest frequency remains unchanged when $r=0$; as $r$ increases, the low frequency is scaled more strongly. It has a smoother transition between local accuracy and long-range coverage than uniform PI. It should be noted that NTK-aware originally came from the open source community experience and was later organized by the YaRN paper system, rather than the independent peer-reviewed paper \cite{en:peng2024yarn} from the beginning.

\subsection{Dynamic NTK: Dynamically select scaling by current sequence length}

Scaling with fixed target length $L_1$ causes short inputs to also suffer positional compression. Dynamic Scaling updates the magnification \cite{en:peng2024yarn} based on the current sequence length $\ell$:
\begin{equation}
  s(\ell)=\max\left(1,\frac{\ell}{L_0}\right),
  \label{en:eq:dynamic-scale}
\end{equation}
Then substitute $s(\ell)$ into ~\eqref{en:eq:ntk-base}. The short sequence maintains the original RoPE and gradually expands after exceeding $L_0$.

This approach conflicts with the KV Cache after rotation: when $s(\ell)$ changes, the rotation angle of the same historical token also changes. If the cached Key has been rotated, the old cache will be inconsistent with the new magnification. Strict implementations either cache the Key before rotation and re-rotate after the magnification changes, or fix the magnification within a generation session; the YaRN paper also clearly reminds this point \cite{en:peng2024yarn}.

\subsection{NTK-by-parts: Select interpolation strategy by frequency interval}

YaRN measures the training coverage of frequency $r$ by the number of rotations within the original training length:
\begin{equation}
  \rho(r)=\frac{L_0\theta_r}{2\pi}.
  \label{en:eq:rotation-count}
\end{equation}
If $\rho(r)$ is very small, it means that the dimension has not been completely rotated within the training interval, and should be fully interpolated like PI to avoid OOD; if $\rho(r)$ is very large, it means that it has been rotated many times, and should be preserved as much as possible to maintain local resolution. Define piecewise linear ramp:
\begin{equation}
  \gamma(\rho)=
  \begin{cases}
    0,&\rho<\alpha,\\
    \dfrac{\rho-\alpha}{\beta-\alpha},&\alpha\le\rho\le\beta,\\
    1,&\rho>\beta,
  \end{cases}
  \label{en:eq:ramp}
\end{equation}
Then the new frequency of NTK-by-parts can be written as:
\begin{equation}
  \theta'_r=
  \bigl(1-\gamma(\rho(r))\bigr)\frac{\theta_r}{s}
  +\gamma(\rho(r))\theta_r.
  \label{en:eq:ntk-by-parts}
\end{equation}
The low-turn frequency is close to PI, the high-turn frequency remains at the original value, and the intermediate frequency transitions smoothly. The empirical setting given by YaRN for the LLaMA series is $\alpha=1,\beta=32$, but this is not the cross-model universal constant \cite{en:peng2024yarn}.

\subsection{YaRN: NTK-by-parts plus attention scale calibration}

Attention entropy often changes after expansion. YaRN adds attention scaling\cite{en:peng2024yarn} based on NTK-by-parts:
\begin{equation}
  \alpha_{mn}=\softmax_n\!\left(
  \frac{q_m^\top k_n}{t\sqrt{d_h}}
  \right),
  \label{en:eq:yarn-temperature}
\end{equation}
And it can be equivalently achieved by scaling $q,k$ at the same time:
\begin{equation}
  \tilde q=\frac{q}{\sqrt{t}},\qquad
  \tilde k=\frac{k}{\sqrt{t}}.
  \label{en:eq:yarn-qk-scale}
\end{equation}
The paper gives an empirical relationship for LLaMA/Llama 2:
\begin{equation}
  \frac{1}{\sqrt{t}}=0.1\ln s+1.
  \label{en:eq:yarn-fit}
\end{equation}
So YaRN is not "another single interpolation formula";\term{frequency division interpolation$+$Attention temperature calibration}. The paper reports that it uses about less than PI$2.5\times$The number of training steps and data achieve similar expansion effects\cite{en:peng2024yarn}. in actual framework\textnormal{factor}, \textnormal{beta\_fast}, \textnormal{beta\_slow}, \textnormal{attention\_factor}The naming may be different and must be consistent with the checkpoint training configuration.

\subsection{LongRoPE: Per-Dimension and Position-Dependent Scaling Search}

The aforementioned methods all use parsing rules to generate scaling. LongRoPE further exploits two types of non-uniformity \cite{en:ding2024longrope}:
\begin{enumerate}
\item Different RoPE dimensions require different scaling factors $\lambda_r$;
\item The position at the beginning of the sequence is more sensitive to interpolation, and the retention interval $n_{\mathrm{hat}}$ can be set.
\end{enumerate}
Its generalized frequency is written as:
\begin{equation}
  \theta'_r=\frac{\theta_r}{\lambda_r},
  \label{en:eq:longrope-scale}
\end{equation}
And use evolutionary search to find $\{\lambda_r\}$ and $n_{\mathrm{hat}}$ in the constraint space. LongRoPE first adapted at 256K length, and then performed second-stage search and interpolation based on this model. The paper reported that it reached 2048K; at the same time, it searched for another set of scaling configurations for the short context to restore the original length performance \cite{en:ding2024longrope}. The key idea is: \term{ expansion magnification should not be assumed to be uniform across dimensions and locations. }

\subsection{LongRoPE2: From Theoretical Periods to Effective Training Periods}

LongRoPE2 pointed out that low-frequency/high-dimensional RoPE not only has a long theoretical period in the pre-training corpus, but also has few samples that actually have long-distance dependence, resulting in its\term{Effective phase range}Narrower than theoretical predictions\cite{en:longrope2}. Therefore, the theoretical critical dimension may not be equal to the actual critical dimension. It makes three improvements:
\begin{enumerate}
\item uses needle-driven perplexity to guide evolutionary search, jointly finding the actual critical dimensions and dimension-by-dimension scaling factors;
\item imposes a monotonic non-decreasing $\lambda_r$ constraint on the OOD dimension;
\item Mixed context training: short samples use original RoPE, long samples use rescaled RoPE.
\end{enumerate}
The hybrid training goal can be abstracted as:
\begin{equation}
  \mathcal{L}=\eta\,\mathcal{L}_{\mathrm{short}}(\theta_{\mathrm{orig}})
  +(1-\eta)\,\mathcal{L}_{\mathrm{long}}(\theta_{\mathrm{scaled}}),
  \label{en:eq:mixed-context-loss}
\end{equation}
It directly handles the conflict between "long context adaptation" and "short context preservation". The paper reports 128K effective context and higher short context retention rate on LLaMA3-8B and Phi3-mini; these numbers are the experimental conclusions of the paper and should not be extrapolated to the default guarantee \cite{en:longrope2} for all models.

\subsection{Method comparison and applicable conditions}

\begin{table}[htbp]
\centering
\small
\begin{tabularx}{\linewidth}{@{}p{2.4cm}p{3.0cm}p{3.2cm}X@{}}
\toprule
Method & Scale object & Main benefits & Main costs/risks \\
\midrule
PI & All positions/frequencies are uniformly divided by $s$ & Stable and simplest to implement & High-frequency local resolution decreases, usually fine-tuning is required \\
NTK-aware & Modifies base, dimensionally non-uniform scaling & retains the highest frequency local structure & Originates from community experience, super parameters are based on model \\
Dynamic NTK & The magnification changes with the current length & Short input keeps the original RoPE & Conflicts with KV Cache after rotation \\
NTK-by-parts & Partition interpolation according to the number of rotations & High frequency retention, low frequency protection OOD & Need to select $\alpha,\beta$ \\
YaRN & NTK-by-parts + temperature calibration & Long and short range compromise is more complete & Must match training recipe and attention factor \\
LongRoPE & Searches for dimension-wise factors and position retention areas & Can exploit non-uniformity and support progressive expansion & Higher search and verification costs \\
LongRoPE2 & Search for effective critical dimensions + hybrid training & Optimize effective length and short-range preservation simultaneously & Requires specialized data, search and mid-training \\
\bottomrule
\end{tabularx}
\caption{A unified comparison of the main RoPE extension methods.}
\label{en:tab:rope-scaling}
\end{table}

\noteBox{\textbf{Practical suggestions: } For existing checkpoints, the native RoPE configuration and training recipe should be reused; \textnormal{rope\_theta}, PI factor, YaRN factor, and LongRoPE's dimension-wise factors are not equivalent. When training a long context model from scratch, RoPE scaling, long and short sample mix, position-id semantics and evaluation matrix need to be jointly designed. }

\section{An Alternative Design: ALiBi's Linear Distance Bias}

ALiBi does not rotate the vector, but adds a linear distance penalty \cite{en:press2022} to the attention score of the $h$ head:
\begin{equation}
  s_{ij}^{(h)}=
  \frac{q_i^{(h)\top}k_j^{(h)}}{\sqrt{d_h}}
  -m_h|i-j|.
  \label{en:eq:alibi}
\end{equation}
Different heads use different slopes $m_h$. It is simple in form, has no position table, and explicitly injects recency bias; but its inductive bias is different from the multi-frequency phase of RoPE and is a relative position bias branch rather than an extension of RoPE.

\section{A unified perspective on three types of positional mechanisms}

Three types of methods modify the input, token pair score and Query-Key geometric relationship respectively:
\begin{align}
x_i'&=x_i+p_i, &&\text{ absolute position }, \label{en:eq:unified-absolute}\\
s_{ij}&=\frac{q_i^\top k_j}{\sqrt{d_h}}+b(i-j), &&\text{ relative position bias}, \label{en:eq:unified-relative}\\
  s_{ij}&=\frac{(R_iq_i)^\top(R_jk_j)}{\sqrt{d_h}}
  =\frac{q_i^\top R_{j-i}k_j}{\sqrt{d_h}}, &&\text{RoPE}. \label{en:eq:unified-rope}
\end{align}

\begin{table}[htbp]
\centering
\footnotesize
\begin{tabularx}{\linewidth}{@{}p{2.2cm}p{1.5cm}p{1.8cm}p{1.8cm}X@{}}
\toprule
Mechanism & Injection position & Relative distance & Extra long position can be calculated & Reliable extrapolation conditions \\
\midrule
Learned Absolute & Input & Not explicit & Normally no & Limited by position table and training length \\
Sinusoidal & Input & Implied & in geometry is & Computable does not equal exploitable \\
Relative Bias & logits/K/V & Explicit & Depends on bucketing & Depends on bucketing and training \\
RoPE & Q/K & Explicit & in dot product Yes & Usually scaling and adaptation training is required \\
ALiBi & logits & explicit & is & with strong recency bias \\
\bottomrule
\end{tabularx}
\caption{A structural comparison of positional mechanisms.}
\label{en:tab:position-compare}
\end{table}

\section{Engineering Experimentation and Evaluation Recommendations}

\subsection{Position Configuration Must Be Bound to the Checkpoint}

For decoder-only LLMs, at least save and verify:
\begin{itemize}
\item \textnormal{rope\_theta} or base frequency;
\item rotary dimension: full dimension or partial rotary;
\item dimension pairing layout;
\item original training length and target length;
\item scaling type, factor, attention factor and dimension-wise parameters;
Definition of position id under \item packing, prefix cache, and sliding window.
\end{itemize}

\subsection{Long-Context Evaluation Must Go Beyond Input Acceptance}

It is recommended to report at least:
\begin{enumerate}
\item short context retention rate: original length PPL and benchmark before and after expansion;
\item PPL bucketed by token position and dependent span;
\item needle/passkey length -- depth two-dimensional grid;
\item RULER class multi-task long context evaluation instead of single retrieval;
\item Valid context: The point where performance starts to degrade significantly, rather than configuring the maximum length;
\item prefill/decode latency, peak memory, KV Cache occupancy and throughput.
\end{enumerate}

The code model should also supplement definition-use span, cross-file import/call graph positioning, same-name symbol disambiguation, repo-level completion and patch correctness, and verify whether the short-range syntax and algorithmic capabilities of the HumanEval class are degraded after the extension.

\section{Common misunderstandings}

\begin{enumerate}
\item \textbf{Self-Attention The order is completely invisible. } More precisely, Self-Attention without position mechanism is equivariant to input displacement. The causal mask provides visible directions but is not equivalent to a fine distance representation.
\item \textbf{ sine code is an absolute position, so it does not contain relative information. The formula } ~\eqref{en:eq:sin-shift} and the formula ~\eqref{en:eq:sin-dot} indicate that the fixed offset corresponds to the fixed rotation, and the dot product only depends on $m-n$.
\item \textbf{RoPE is a relative position encoding, so it does not encode an absolute position. } single $R_mq_m$ explicitly depends on $m$; the absolute phases cancel only in the dot product.
\item \textbf{ position function can be calculated to 128K, and the model has 128K context. } configuration length, value runnability length and effective context length are three things.
\item \textbf{RoPE naturally causes attention to decrease monotonically with distance. } formula ~\eqref{en:eq:rope-oscillation} is a periodic function, and there is no strict monotonic guarantee on a sample-by-sample basis.
\item \textbf{ Just increase \textnormal{rope\_theta} to complete the expansion. } frequency scaling is only the first step. Model adaptation, KV Cache consistency, short-range persistence and valid context must also be verified.
\end{enumerate}

\section{Summary and Outlook}

Absolute position encoding writes coordinates into token representations, relative position encoding directly modifies the relationship between token pairs, and RoPE converts absolute positions into relative phases in dot products through Query-Key rotation. PI, NTK-aware, YaRN and LongRoPE further deal with frequency scaling and long and short context compatibility issues; RoPE scaling law starts from period coverage and links base, training length, critical dimension and extrapolation range. However, there are still several unresolved issues with the existing location mechanism:

\begin{enumerate}
\item \textbf{Nominal length is not effective length.} Accepting 128K or 1M input tokens does not mean that all positions are used reliably. Effective context also depends on the dependency spans represented in training, optimization behavior, and attention dilution.
\item \textbf{ periodically brings phase aliasing. } Different distances may produce similar phases; multi-frequency combination can alleviate, but cannot eliminate this problem theoretically.
\item \textbf{ expansion and short-range capabilities still conflict. } interpolation will reduce local resolution, and dimension-by-dimensional scaling and mixed-length training also lack stable rules that can be transferred across models.
\item \textbf{ system implementation may break location semantics. }packing, prefix cache, sliding window, speculative decoding and KV Cache reuse may cause hidden position-id misalignment.
\item \textbf{ One-dimensional distance is difficult to express complex structures. The file hierarchy, call graph and definition-use relationships of the } code, as well as the spatial and temporal structure of the multimodal data, are beyond the expression range of a single sequence coordinate.
\item \textbf{ position encoding and efficient attention lack a unified design. } Sparse attention, sliding window, KV compression and linear attention will change the information propagation path, and the position mechanism needs to be designed together with the operator.
\item \textbf{ evaluation method is still weak. }needle-in-a-haystack does not adequately reflect multi-hop reasoning, cross-file code understanding, and long-range causal dependencies.
\end{enumerate}

Follow-up research should not only amplify the nominal context, but establish a location mechanism that is predictable, trainable, cacheable, and capable of expressing multi-dimensional structures, and examine long-range information utilization with real long-dependency tasks. For code models, definition-use, cross-file calls, and repository-level modifications are closer to actual needs than random needles.

\clearpage
\appendix
\section{Position-Encoding Choices in Representative LLMs}

Table ~\ref{en:tab:llm-position-encoding} summarizes the position mechanisms that can be verified in public papers, technical reports or official open source configurations. The data status is updated to August 7, 2026. The "context extension" in the table only records the practices related to positional encoding, and does not mean that the model can reach the nominal context length by relying on this mechanism alone. For models that do not disclose architectural details, the positional encoding cannot be deduced based on the context length or old models of the same series.

\begin{longtable}{@{}p{2.7cm}p{3.6cm}p{8.0cm}@{}}
\caption{Publicly disclosed positional encodings for representational language models.}
\label{en:tab:llm-position-encoding}\\
\toprule
Model/series & Position mechanism & Related designs in technical report \\
\midrule
\endfirsthead
\multicolumn{3}{l}{\small Table ~\thetable (continued) }\\
\toprule
Model/series & Position mechanism & Related designs in technical report \\
\midrule
\endhead
\midrule
\multicolumn{3}{r}{\small Continued on next page}\\
\endfoot
\bottomrule
\endlastfoot

GPT-2 / GPT-3 & Learnable absolute position embedding & GPT-2 uses learnable position embedding; GPT-3 follows the basic Transformer architecture of GPT-2, so it still uses the learned absolute solution \cite{en:radford2019gpt2,en:brown2020gpt3}. The maximum length for this type of design is usually constrained by the location table and training length. \\

T5 & bucket relative position bias & Each attention head learns a scalar bias for the relative distance bucket; short-distance subdivision, long-distance approximate logarithmic bucket \cite{en:raffel2020}. \\

BLOOM & ALiBi & does not add position vectors to the embedding, but instead applies a linear negative bias \cite{en:lescao2022bloom} that grows with distance on the attention logits. \\

Falcon & ALiBi & Falcon Series technical report using ALiBi; ablation experiments comparing ALiBi, URPE and RoPE\cite{en:almazrouei2023falcon} simultaneously. \\

PaLM & RoPE & Use RoPE instead of absolute or relative position embedding; the report lists its long sequence performance as one of the reasons for adoption \cite{en:chowdhery2022palm}. \\

LLaMA / Llama 2 & RoPE & LLaMA replaces absolute position embedding with RoPE; Llama 2 continues RoPE and expands the pre-training context from 2K to 4K\cite{en:touvron2023llama,en:touvron2023llama2}. \\

Code Llama & RoPE, increase the base &. Based on Llama 2, adjust the RoPE base to $10^6$, and continue training using 16K sequences; the report studies the extrapolation of \cite{en:roziere2023codellama} to 100K. \\

Llama 3 & RoPE, $\theta=500{,}000$ & technical report increases RoPE base to 500K to improve long context support; this configuration is consistent with \cite{en:dubey2024llama3} for 8B, 70B and 405B models. \\

Qwen2 & RoPE + YaRN + DCA & In the later stage of pre-training, the context is expanded from 4K to 32K, and the RoPE base is adjusted from 10K to $10^6$; the inference side combines YaRN and Dual Chunk Attention to support longer input \cite{en:yang2024qwen2}. \\

Qwen3 / Qwen3.6 & RoPE Series + YaRN & The official configuration of Qwen3 uses $\theta=10^6$, scalable from native 32K to 131K\cite{en:yang2025qwen3} via YaRN. Qwen3.6 further adopts a hybrid structure of full attention and linear attention; its official configuration uses partially rotated M-RoPE, $\theta=10^7$, in the full attention layer, natively supports 262K, and provides a configuration \cite{en:qwen36config} that uses YaRN to expand to about 1M. \\

Gemma & RoPE & Gemma uses rotary positional embeddings at each layer instead of absolute positional embedding; the basic model training length is 8K\cite{en:gemmateam2024}. \\

Phi-3 / Phi-3.5 & RoPE system architecture + LongRoPE & Phi-3-mini uses a block similar to Llama 2; the 128K version is extended by LongRoPE, and Phi-3.5 continues to use LongRoPE and mixed context training \cite{en:abdin2024phi3}. \\

The decoupled RoPE & in DeepSeek-V2 / V3 & MLA is compatible with low-rank KV compression and decouples the Query/Key subspace carrying RoPE from the content compression subspace; V3 continues this MLA design \cite{en:deepseek2024v2,en:deepseek2024v3}. \\

The decoupled RoPE & LongCat-Flash, LongCat-Flash-Thinking and 2601 versions of LongCat series & MLA follow MLA and divide Query/Key into RoPE and NoPE subspaces. The official configuration of the first version of Flash is 64-dimensional RoPE subspace, 128-dimensional NoPE subspace, $\theta=10^7$, and the maximum position is 131K; the public configuration of 2601 sets $\theta$ to $10^6$\cite{en:longcatflash,en:longcatthinking2601,en:longcatconfigs}. \\

Kimi K3 & NoPE + Kimi Delta Attention & does not inject explicit position encoding into MLA's Query/Key; position sensitivity is provided implicitly by KDA's recursive gating and decay mechanism. The technical report states that the design can continue training to 1M context \cite{en:kimik3} without RoPE rescaling or interpolation. \\

GLM-5 & decoupled RoPE & in MLA/DSA GLM-5 is calculated with the attention of DSA sparse MLA; the official configuration retains the 64-dimensional RoPE subspace and the 192-dimensional NoPE subspace, and sets $\theta=10^6$. DSA changes sparse retrieval path without canceling RoPE\cite{en:glm5,en:glm5config}. \\

Qwen3.8-Max & Unpublished & Official service documentation lists Qwen3.8-Max, but as of the date of this article, there are no public technical reports, weights, or configurations to confirm the location mechanism. Therefore, the RoPE solution \cite{en:qwen38docs} of Qwen3/Qwen3.6 cannot be directly applied. \\

GPT-4 & Unpublished & The GPT-4 technical report does not disclose the model structure, position encoding or RoPE configuration, so its position mechanism \cite{en:openai2023gpt4} cannot be determined from public materials. \\

\end{longtable}

It can be seen from public information that decoder-only open source models have obviously concentrated on RoPE after 2022, but they have not fully converged: BLOOM and Falcon use ALiBi, the T5 series model uses relative position bias, DeepSeek, LongCat and GLM-5 split the RoPE/NoPE subspace to adapt to MLA or sparse attention, and Kimi K3 turns to the NoPE scheme in which KDA implicitly provides position information. The so-called "use RoPE" is not a complete configuration. Base, rotary dimension, scaling, training length and position-id semantics will all change the actual behavior; for models with undisclosed architectures such as Qwen3.8-Max, keeping "unknown" in the table is more reliable than following the series of historical configurations.

\clearpage
\pdfbookmark[0]{中文版}{bilingual.chinese}
\setcounter{section}{0}
\setcounter{subsection}{0}
\setcounter{subsubsection}{0}
\setcounter{equation}{0}
\setcounter{figure}{0}
\setcounter{table}{0}
\setcounter{footnote}{0}
\setcounter{tocdepth}{2}
\renewcommand{\abstractname}{摘要}
\renewcommand{\contentsname}{目录}
\renewcommand{\figurename}{图}
\renewcommand{\tablename}{表}
\renewcommand{\refname}{参考文献}
\renewcommand{\appendixname}{附录}
\BilingualUseChineseToc
\renewcommand*{\theHsection}{zh.\arabic{section}}
\renewcommand*{\theHsubsection}{zh.\arabic{section}.\arabic{subsection}}
\renewcommand*{\theHsubsubsection}{zh.\arabic{section}.\arabic{subsection}.\arabic{subsubsection}}
\renewcommand*{\theHequation}{zh.\arabic{equation}}
\renewcommand*{\theHfigure}{zh.\arabic{figure}}
\renewcommand*{\theHtable}{zh.\arabic{table}}
\gdef\BilingualAnchorPrefix{zh}
\fancyhead[L]{\small\color{gray}Transformer 位置编码的前世今生}
\markboth{}{}
\begin{center}
  {\zihao{1}\bfseries\color{navy} Transformer 位置编码的前世今生}\par
  \vspace{4pt}
  {\zihao{3}\color{navy} 从绝对位置、相对位置到旋转位置编码 RoPE}\par
  \vspace{10pt}
  {\normalsize\textbf{Jiguo Li\footnote{本文在Codex协助下完成}}}\par
  \vspace{2pt}
  {\small\href{mailto:jiguolee@gmail.com}{jiguolee@gmail.com}}
\end{center}
\vspace{5pt}
\hrule

\begin{abstract}
Self-Attention 能有效建模 token 之间的内容相关性，但其计算本身不包含序列顺序。位置编码的作用，是把绝对坐标、相对距离或旋转相位注入注意力，使模型能够区分词序、表达局部结构，并在更长上下文中保持可用的位置信号。

本文从 Self-Attention 的置换等变性出发，依次推导正弦--余弦位置编码、Shaw 相对位置表示、Transformer-XL、T5 Relative Position Bias 与 RoPE，比较它们注入位置的环节、计算代价、长度外推能力及对 KV Cache 的影响。在此基础上，进一步梳理 Position Interpolation、RoPE scaling law、NTK-aware、Dynamic NTK、NTK-by-parts、YaRN、LongRoPE 和 LongRoPE2，说明这些方法如何调整频率、位置尺度与注意力温度，以及训练长度、目标上下文长度和微调数据之间的约束。全文同时给出公式推导、直观解释、实现要点与一手文献，便于在模型设计和长上下文扩展中据此选择方案。
\end{abstract}

\noindent\textbf{关键词：}Transformer；位置编码；相对位置编码；RoPE；长上下文；长度外推

\noteBox{\textbf{阅读提要：}绝对位置编码为 token 指定坐标；相对位置编码描述 token 对之间的距离；RoPE 则把位置写入 Query/Key 的旋转相位。}

\clearpage
\begingroup
\small
\setlength{\parskip}{0pt}
\linespread{1.02}\selectfont
\setcounter{tocdepth}{2}
\tableofcontents
\endgroup
\clearpage

\section{为什么 Transformer 必须显式注入位置}

\subsection{RNN、CNN 与 Transformer 的差异}

RNN 按时间递推：
\begin{equation}
  h_t=f(x_t,h_{t-1}),
  \label{zh:eq:rnn}
\end{equation}
计算路径本身就是 $x_1\rightarrow x_2\rightarrow\cdots\rightarrow x_n$，顺序天然存在于状态传递中。CNN 的卷积核先聚合局部邻域，因此也带有局部结构先验。

Transformer 去掉递归和卷积，让所有 token 并行交互。这是高吞吐的来源，也意味着模型结构本身不再知道 token 的先后顺序。原始 Transformer 因此必须额外注入位置信息\cite{zh:vaswani2017}。

\subsection{不带位置的 Self-Attention 看到了什么}

给定输入矩阵：
\begin{equation}
  X=[x_1,x_2,\ldots,x_n]^\top\in\R^{n\times d},
  \label{zh:eq:input}
\end{equation}
标准 Self-Attention 计算：
\begin{align}
  Q&=XW_Q,\quad K=XW_K,\quad V=XW_V, \label{zh:eq:qkv}\\
  \Attn(X)&=\softmax\!\left(\frac{QK^\top}{\sqrt{d_h}}\right)V. \label{zh:eq:attention}
\end{align}
第 $i$ 个 Query 对第 $j$ 个 Key 的分数为：
\begin{equation}
  s_{ij}=\frac{q_i^\top k_j}{\sqrt{d_h}}.
  \label{zh:eq:score}
\end{equation}
该分数使用了内容向量，却没有直接使用位置 $i,j$。设 $P$ 为任意置换矩阵，则：
\begin{equation}
  \Attn(PX)=P\Attn(X).
  \label{zh:eq:permutation}
\end{equation}
因此，不含位置机制的 Self-Attention 是\term{置换等变}的：打乱输入后，输出只按相同方式打乱。位置可以在输入表示、注意力分数或 Query/Key 的几何变换中注入，分别对应绝对位置、相对位置和 RoPE 三类方案。

\section{技术演进概览}

\begin{table}[htbp]
\centering
\small
\begin{tabularx}{\linewidth}{@{}p{1.2cm}p{3.5cm}X@{}}
\toprule
时间 & 代表工作 & 核心位置机制 \\
\midrule
2017 & Transformer & 固定正弦--余弦绝对位置编码 \\
2018 & BERT & 可学习绝对位置 embedding \\
2018 & Shaw et al. & 在注意力中加入相对距离表示 \\
2019 & Transformer-XL & 相对位置分解，支持片段级记忆 \\
2020 & T5 & 每个注意力头学习分桶距离 bias \\
2021 & RoFormer / RoPE & 按位置旋转 Query 和 Key \\
2021 & ALiBi & 对 logits 加线性距离惩罚 \\
2023 & PI / YaRN & RoPE 插值、分频缩放与注意力校准 \\
2024--2025 & LongRoPE / LongRoPE2 & 搜索逐维缩放因子并混合长短上下文训练 \\
\bottomrule
\end{tabularx}
\caption{Transformer 位置机制的代表性演进。}
\label{zh:tab:timeline}
\end{table}

这些方法不是简单的新旧替代。固定长度 encoder、encoder-decoder 与自回归 LLM 的任务结构、推理方式和上下文需求不同，因此多种方案至今并存。

\section{第一代：绝对位置编码}

绝对位置编码为位置 $i$ 提供向量 $p_i$：
\begin{equation}
  h_i^{(0)}=e(x_i)+p_i.
  \label{zh:eq:absolute}
\end{equation}
它相当于给每个 token 贴上“第 137 位”这样的全局坐标。

\subsection{可学习绝对位置 embedding}

最直接的实现是维护参数表：
\begin{equation}
  P\in\R^{L_{\max}\times d}.
  \label{zh:eq:learned-table}
\end{equation}
第 $i$ 行就是位置 $i$ 的向量 $p_i$。BERT 将 token、segment 与 position embedding 相加作为输入\cite{zh:devlin2019}。它实现简单、可直接拟合任务内位置模式，但参数表绑定最大长度；未训练位置没有可靠表示；相邻位置差不被结构性约束；内容与位置在第一层前即混合：
\begin{equation}
  q_i=(e(x_i)+p_i)W_Q=e(x_i)W_Q+p_iW_Q.
  \label{zh:eq:absolute-proj}
\end{equation}

\subsection{正弦--余弦位置编码}

原始 Transformer 用多组正弦和余弦函数生成位置向量\cite{zh:vaswani2017}：
\begin{align}
  PE(pos,2i)&=\sin\!\left(\frac{pos}{10000^{2i/d_{\text{model}}}}\right), \label{zh:eq:sin-pe}\\
  PE(pos,2i+1)&=\cos\!\left(\frac{pos}{10000^{2i/d_{\text{model}}}}\right). \label{zh:eq:cos-pe}
\end{align}
定义第 $i$ 组角频率：
\begin{equation}
  \theta_i=10000^{-2i/d_{\text{model}}},
  \label{zh:eq:sin-theta}
\end{equation}
则一组二维表示为：
\begin{equation}
  PE_i(pos)=
  \begin{bmatrix}
    \sin(pos\theta_i)\\
    \cos(pos\theta_i)
  \end{bmatrix}.
  \label{zh:eq:sin-pair}
\end{equation}
完整位置向量由多组不同频率拼接：
\begin{equation}
  PE(pos)=[\sin(pos\theta_0),\cos(pos\theta_0),\ldots,
  \sin(pos\theta_{d/2-1}),\cos(pos\theta_{d/2-1})].
  \label{zh:eq:sin-full}
\end{equation}

\subsubsection{“多只钟表”的直观解释}

每对维度可看成单位圆上的一根指针，其相位与波长分别为：
\begin{align}
  \phi_i(pos)&=pos\theta_i, \label{zh:eq:phase}\\
  \lambda_i&=\frac{2\pi}{\theta_i}=2\pi\cdot10000^{2i/d_{\text{model}}}. \label{zh:eq:wavelength}
\end{align}
大的 $\theta_i$ 对应“快钟”，能区分相邻位置；小的 $\theta_i$ 对应“慢钟”，描述较长尺度。单只钟周期重复，多只不同转速的钟联合后能在更大范围区分位置。常数 $10000$ 是频率基数，不是数学上唯一正确的选择。

当 $d_{\text{model}}=8$ 时，四组频率为：
\begin{equation}
  \theta_0=1,\qquad \theta_1=0.1,\qquad
  \theta_2=0.01,\qquad \theta_3=0.001.
  \label{zh:eq:d8-frequency}
\end{equation}

\subsubsection{为什么必须同时使用 sin 和 cos}

由和角公式：
\begin{equation}
  \sin((pos+\Delta)\theta)=
  \sin(pos\theta)\cos(\Delta\theta)+
  \cos(pos\theta)\sin(\Delta\theta).
  \label{zh:eq:addition}
\end{equation}
只保存 sin 无法通过固定线性变换得到平移结果；成对保存后有：
\begin{equation}
  PE_i(pos+\Delta)=
  \begin{bmatrix}
    \cos(\Delta\theta_i)&\sin(\Delta\theta_i)\\
    -\sin(\Delta\theta_i)&\cos(\Delta\theta_i)
  \end{bmatrix}PE_i(pos).
  \label{zh:eq:sin-shift}
\end{equation}
矩阵只依赖位移 $\Delta$。同一频率下两个位置编码的点积为：
\begin{equation}
  PE_i(m)^\top PE_i(n)=\cos((m-n)\theta_i),
  \label{zh:eq:sin-dot}
\end{equation}
完整编码满足：
\begin{equation}
  PE(m)^\top PE(n)=\sum_i\cos((m-n)\theta_i).
  \label{zh:eq:sin-full-dot}
\end{equation}
所以正弦编码形式上是绝对位置表示，几何关系中却包含相对位移。原始 Transformer 实际使用式~\eqref{zh:eq:absolute}，并未直接以式~\eqref{zh:eq:sin-full-dot} 作为注意力分数，模型仍需训练才能利用这种结构。

\noteBox{\textbf{需要注意：}位置函数可以计算到任意 $pos$，不等于整个模型能可靠理解任意长上下文。训练长度仍约束注意力分布、状态表示与优化结果。}

\section{第二代：相对位置编码}

语言和代码中的许多关系更关心 $j-i$，而不是分别知道 $i$ 与 $j$。例如“左侧最近的定义”“距调用点最近的声明”“右侧两个 token 的参数”。

\subsection{Shaw：把相对距离加入 Key 和 Value}

Shaw 等人将 token 对的相对距离直接加入 Self-Attention\cite{zh:shaw2018}。先裁剪距离：
\begin{equation}
  r_{ij}=\operatorname{clip}(j-i,-K,K),
  \label{zh:eq:clip}
\end{equation}
为每个距离学习相对 Key 向量 $a^K_{r_{ij}}$：
\begin{equation}
  s_{ij}=\frac{q_i^\top(k_j+a^K_{r_{ij}})}{\sqrt{d_h}}
  =\frac{q_i^\top k_j+q_i^\top a^K_{r_{ij}}}{\sqrt{d_h}}.
  \label{zh:eq:shaw-score}
\end{equation}
第一项表示内容匹配，第二项表示当前 Query 对某个方向与距离的偏好。Value 也可加入相对表示：
\begin{equation}
  z_i=\sum_j\alpha_{ij}(v_j+a^V_{r_{ij}}).
  \label{zh:eq:shaw-value}
\end{equation}

\subsection{Transformer-XL：跨片段记忆中的相对位置}

Transformer-XL 允许当前片段复用上一片段隐藏状态\cite{zh:dai2019}。若缓存携带旧片段的绝对坐标，复用到新片段会产生歧义；相对位置只依赖当前 Query 与历史 Key 的真实距离。其未归一化注意力可分解为：
\begin{equation}
  \begin{aligned}
  A_{ij}={}&q_i^\top k_j+q_i^\top W_{k,R}R_{i-j}\\
           &+u^\top k_j+v^\top W_{k,R}R_{i-j},
  \end{aligned}
  \label{zh:eq:txl}
\end{equation}
分别对应内容--内容、内容--位置、全局内容偏置和全局位置偏置。

\subsection{T5：将相对位置压缩为标量 bias}

T5 为每个注意力头和距离桶学习一个标量\cite{zh:raffel2020}：
\begin{equation}
  s_{ij}^{(h)}=
  \frac{q_i^{(h)\top}k_j^{(h)}}{\sqrt{d_h}}
  +b_{h,\operatorname{bucket}(j-i)}.
  \label{zh:eq:t5-bias}
\end{equation}
距离较小时使用细粒度桶，距离较大时近似对数分桶。不同 head 可学习不同距离先验，形成“内容相似度 $+$ 距离偏好”的清晰分解。

传统相对位置方法在建模上直观，但通常要处理 $n\times n$ 个 token 对。相对信息可能进入 Key、Value、logits 或多个交叉项，增加算子融合、增量解码与高效注意力实现的复杂度。RoPE 的关键价值，就是在显式表达相对位移的同时保留标准 $QK^\top$ 形式。

\section{第三代：旋转位置编码 RoPE}

RoPE 不把位置向量加到输入，也不为 token 对显式构造相对向量，而是根据绝对位置旋转 Query 和 Key\cite{zh:su2021}。

\subsection{从二维旋转推导核心公式}

二维旋转矩阵为：
\begin{equation}
  R(\phi)=
  \begin{bmatrix}
    \cos\phi&-\sin\phi\\
    \sin\phi&\cos\phi
  \end{bmatrix}.
  \label{zh:eq:rotation}
\end{equation}
位置 $m$ 的旋转角为 $m\theta$：
\begin{equation}
  q'_m=R(m\theta)q_m,\qquad k'_n=R(n\theta)k_n.
  \label{zh:eq:rope-rotate}
\end{equation}
旋转后的点积为：
\begin{equation}
  \begin{aligned}
  q_m'^\top k_n'
  &=q_m^\top R(m\theta)^\top R(n\theta)k_n\\
  &=q_m^\top R((n-m)\theta)k_n.
  \end{aligned}
  \label{zh:eq:rope-core}
\end{equation}
这里使用了：
\begin{equation}
  R(m\theta)^\top=R(-m\theta),\qquad
  R(-m\theta)R(n\theta)=R((n-m)\theta).
  \label{zh:eq:rotation-group}
\end{equation}
单个向量 $q'_m$ 依赖绝对位置 $m$；Query-Key 交互只通过 $n-m$ 依赖相对位置。因此，\term{RoPE 在单个向量上编码绝对位置，在点积中呈现相对位置}。

\subsection{高维 RoPE 与复数形式}

将 head dimension 两两组成二维子空间，第 $r$ 对频率为：
\begin{equation}
  \theta_r=b^{-2r/d_h},\qquad r=0,1,\ldots,d_h/2-1,
  \label{zh:eq:rope-frequency}
\end{equation}
经典基数 $b=10000$。整体旋转矩阵为：
\begin{equation}
  R_m=\operatorname{diag}\left(
  R(m\theta_0),R(m\theta_1),\ldots,R(m\theta_{d_h/2-1})
  \right).
  \label{zh:eq:rope-block}
\end{equation}
若把二维向量写成复数 $z_r=x_{2r}+\mathrm{i}x_{2r+1}$，RoPE 等价于：
\begin{equation}
  z'_r=z_r e^{\mathrm{i}m\theta_r}.
  \label{zh:eq:rope-complex}
\end{equation}
Query 与 Key 的相位因子相乘后为：
\begin{equation}
  e^{-\mathrm{i}m\theta_r}e^{\mathrm{i}n\theta_r}
  =e^{\mathrm{i}(n-m)\theta_r},
  \label{zh:eq:rope-relative-phase}
\end{equation}
即绝对相位相减后只剩相对相位。

\subsection{工程实现与配对约定}

以下实现使用相邻维度配对。某些框架采用前半维与后半维配对；两者只是排列约定不同，但权重、cos/sin cache 与旋转函数必须一致。

\begin{lstlisting}
def apply_rope(x, cos, sin):
    x_even = x[..., 0::2]
    x_odd  = x[..., 1::2]
    y_even = x_even * cos - x_odd * sin
    y_odd  = x_even * sin + x_odd * cos

    y = torch.empty_like(x)
    y[..., 0::2] = y_even
    y[..., 1::2] = y_odd
    return y
\end{lstlisting}

Value 通常不旋转，因为位置主要影响“关注谁”的权重；聚合对象仍是内容 Value。

\subsection{RoPE 与 KV Cache}

固定 RoPE 配置下，历史 Key 通常以旋转后的形式缓存：
\begin{equation}
  k'_i=R_i k_i.
  \label{zh:eq:cached-key}
\end{equation}
生成位置 $t$ 时只需计算：
\begin{equation}
  q'_t=R_tq_t,\qquad k'_t=R_tk_t.
  \label{zh:eq:decode-rope}
\end{equation}
历史 Key 无需重算。工程上最易出错的是 position offset：prefill、decode、padding、sequence packing、prefix cache 与 sliding window 必须对同一 token 使用一致的逻辑 position id。

\subsection{RoPE 是否保证距离越远注意力越弱}

单个频率的相对项包含：
\begin{equation}
  \cos((n-m)\theta_r),\qquad \sin((n-m)\theta_r),
  \label{zh:eq:rope-oscillation}
\end{equation}
它们随距离周期振荡，并非单调衰减。RoFormer 讨论了多频率汇总后的远距离衰减倾向\cite{zh:su2021}，但不能对任意固定 Query/Key、任意 head 和任意距离推出“越远分数必然越小”。

\section{RoPE 长上下文扩展：从统一缩放到逐维搜索}

\subsection{问题根源：外推引入未训练相位}

设原始训练长度为 $L_0$，目标长度为 $L_1=sL_0$，其中 $s>1$。第 $r$ 个二维子空间的周期为：
\begin{equation}
  T_r=\frac{2\pi}{\theta_r}=2\pi b^{2r/d_h}.
  \label{zh:eq:rope-period}
\end{equation}
高频维度在 $L_0$ 内经历许多周期，局部模式训练充分；低频维度可能连一个完整周期都未覆盖。直接推理到 $L_1$ 会让部分维度进入未见过的相位区间，造成 RoPE OOD。LongRoPE2 将理论临界维度定义为周期是否超过原训练长度的分界\cite{zh:longrope2}：
\begin{equation}
  r_{\mathrm{crit}}=\min\{r:T_r\ge L_0\}.
  \label{zh:eq:critical-dim}
\end{equation}
扩展方法真正需要处理的是各频率的缩放方式，以及模型权重对新相位分布的适应，而不仅是增大 position id 的取值范围。

\subsection{RoPE scaling law：基数、训练长度与外推范围}

Liu 等人从周期覆盖的角度研究了 RoPE 的长度外推，并提出 RoPE-based extrapolation scaling laws\cite{zh:liu2024scalingrope}。这里的“scaling law”描述的是旋转基数、训练长度与外推范围之间的关系，与参数量、数据量和计算量之间的经典模型 scaling law 含义不同。

对原始基数 $10000$，第 $n$ 组频率的周期为：
\begin{equation}
  T_n=\frac{2\pi}{\theta_n}=2\pi\cdot10000^{2n/d_h}.
  \label{zh:eq:scaling-period}
\end{equation}
若预训练长度为 $T_{\mathrm{train}}$，论文把能够在训练区间内至少覆盖完整周期的维度上限写为：
\begin{equation}
  d_{\mathrm{extra}}
  =2\left\lceil
  \frac{d_h}{2}\log_{10000}\!\left(\frac{T_{\mathrm{train}}}{2\pi}\right)
  \right\rceil.
  \label{zh:eq:critical-dimension-scaling-law}
\end{equation}
超过 $d_{\mathrm{extra}}$ 的低频维度在训练中没有观察到完整周期，超出训练长度后更容易出现相位分布偏移。以 LLaMA 2 的 $d_h=128$、$T_{\mathrm{train}}=4096$ 为例，论文计算得到 $d_{\mathrm{extra}}=92$；其余 36 个维度被视为外推时更容易失稳的部分\cite{zh:liu2024scalingrope}。

当微调阶段把旋转基数改为 $\beta>10000$，该工作用临界维度的新周期估计外推上限：
\begin{equation}
  T_{\mathrm{extra}}
  =2\pi\,\beta^{d_{\mathrm{extra}}/d_h}.
  \label{zh:eq:scaling-law-upper-bound}
\end{equation}
反过来，给定期望长度 $\widetilde T_{\mathrm{extra}}$，可由论文给出的关系估计所需的临界基数：
\begin{equation}
  \beta_0
  =10000^{\log_{T_{\mathrm{train}}/(2\pi)}
  \left(\widetilde T_{\mathrm{extra}}/(2\pi)\right)}.
  \label{zh:eq:scaling-law-critical-base}
\end{equation}

论文还观察到，在固定微调长度下，将基数调小也可能改善外推，因为更多维度能在训练区间内经历更充分的三角函数变化。对应的三个相位覆盖点为：
\begin{equation}
  \beta_1=\frac{2T_{\mathrm{train}}}{\pi},\qquad
  \beta_2=\frac{T_{\mathrm{train}}}{\pi},\qquad
  \beta_3=\frac{T_{\mathrm{train}}}{2\pi}.
  \label{zh:eq:scaling-law-small-base}
\end{equation}
这组结果揭示了 RoPE 外推与周期覆盖之间的联系，但不宜把式~\eqref{zh:eq:scaling-law-upper-bound} 当成跨模型的硬性上下文上限。论文主要在 LLaMA 2 上验证，实际有效长度还受到语料中的长距离依赖分布、微调数据、attention head 行为和评测任务影响。

\subsection{Position Interpolation：统一压缩位置索引}

PI 将目标范围中的位置压回原训练区间\cite{zh:chen2023pi}：
\begin{equation}
  m'=\frac{m}{s}=m\frac{L_0}{L_1}.
  \label{zh:eq:pi-position}
\end{equation}
等价地，对所有频率统一缩放：
\begin{equation}
  \theta'_r=\frac{\theta_r}{s}.
  \label{zh:eq:pi-frequency}
\end{equation}
其优点是所有新位置都落回已训练相位范围，结构改动极小；缺点是高频也被压缩，相邻 token 的相位差从 $\theta_r$ 变成 $\theta_r/s$，局部分辨率下降。PI 论文从 2K 扩至 32K，仅需短程微调即可适应，但也明确指出原长度内会有一定性能损失\cite{zh:chen2023pi}。

\subsection{NTK-aware：通过修改 RoPE base 非均匀缩放频率}

NTK-aware 的直觉是：不要像 PI 一样等比例压缩所有频率，而是调整频率基数 $b$，使高频尽量保持、低频承担更多扩展。常见形式把新基数设为：
\begin{equation}
  b'=b\,s^{d_h/(d_h-2)},
  \label{zh:eq:ntk-base}
\end{equation}
从而：
\begin{equation}
  \theta'_r=(b')^{-2r/d_h}
  =\theta_r\,s^{-2r/(d_h-2)}.
  \label{zh:eq:ntk-frequency}
\end{equation}
当 $r=0$ 时最高频保持不变；随 $r$ 增大，低频被更强地缩放。它在局部精度和长程覆盖之间比统一 PI 更平滑。需要注意，NTK-aware 最初来自开源社区经验，后来由 YaRN 论文系统整理，而不是一开始就有独立同行评议论文\cite{zh:peng2024yarn}。

\subsection{Dynamic NTK：按当前序列长度动态选择缩放}

固定目标长度 $L_1$ 的缩放会使短输入也承受位置压缩。Dynamic Scaling 根据当前序列长度 $\ell$ 更新倍率\cite{zh:peng2024yarn}：
\begin{equation}
  s(\ell)=\max\left(1,\frac{\ell}{L_0}\right),
  \label{zh:eq:dynamic-scale}
\end{equation}
再将 $s(\ell)$ 代入式~\eqref{zh:eq:ntk-base}。短序列保持原 RoPE，超过 $L_0$ 后逐步扩展。

这一做法与旋转后 KV Cache 存在冲突：当 $s(\ell)$ 变化时，同一个历史 token 的旋转角也变化。若缓存的是已经旋转的 Key，旧缓存便与新倍率不一致。严格实现要么缓存旋转前 Key 并在倍率变化后重新旋转，要么在一次生成会话内固定倍率；YaRN 论文也明确提醒了这一点\cite{zh:peng2024yarn}。

\subsection{NTK-by-parts：按频率区间选择插值策略}

YaRN 用原训练长度内的旋转圈数衡量第 $r$ 个频率的训练覆盖：
\begin{equation}
  \rho(r)=\frac{L_0\theta_r}{2\pi}.
  \label{zh:eq:rotation-count}
\end{equation}
若 $\rho(r)$ 很小，说明该维度在训练区间内尚未完整旋转，应像 PI 一样充分插值以避免 OOD；若 $\rho(r)$ 很大，说明它已经多次旋转，应尽量保留以维持局部分辨率。定义分段线性 ramp：
\begin{equation}
  \gamma(\rho)=
  \begin{cases}
    0,&\rho<\alpha,\\
    \dfrac{\rho-\alpha}{\beta-\alpha},&\alpha\le\rho\le\beta,\\
    1,&\rho>\beta,
  \end{cases}
  \label{zh:eq:ramp}
\end{equation}
则 NTK-by-parts 的新频率可写为：
\begin{equation}
  \theta'_r=
  \bigl(1-\gamma(\rho(r))\bigr)\frac{\theta_r}{s}
  +\gamma(\rho(r))\theta_r.
  \label{zh:eq:ntk-by-parts}
\end{equation}
低圈数频率接近 PI，高圈数频率保持原值，中间频率平滑过渡。YaRN 对 LLaMA 系列给出的经验设置为 $\alpha=1,\beta=32$，但这不是跨模型通用常数\cite{zh:peng2024yarn}。

\subsection{YaRN：NTK-by-parts 加注意力尺度校准}

扩长后 attention entropy 往往改变。YaRN 在 NTK-by-parts 基础上加入 attention scaling\cite{zh:peng2024yarn}：
\begin{equation}
  \alpha_{mn}=\softmax_n\!\left(
  \frac{q_m^\top k_n}{t\sqrt{d_h}}
  \right),
  \label{zh:eq:yarn-temperature}
\end{equation}
并可通过同时缩放 $q,k$ 来等价实现：
\begin{equation}
  \tilde q=\frac{q}{\sqrt{t}},\qquad
  \tilde k=\frac{k}{\sqrt{t}}.
  \label{zh:eq:yarn-qk-scale}
\end{equation}
论文对 LLaMA/Llama 2 给出经验关系：
\begin{equation}
  \frac{1}{\sqrt{t}}=0.1\ln s+1.
  \label{zh:eq:yarn-fit}
\end{equation}
因此 YaRN 不是“另一种单一插值公式”，而是\term{分频率插值 $+$ 注意力温度校准}。论文报告它用比 PI 少约 $2.5\times$ 的训练步数和数据达到相近扩展效果\cite{zh:peng2024yarn}。实际框架中的 \textnormal{factor}、\textnormal{beta\_fast}、\textnormal{beta\_slow}、\textnormal{attention\_factor} 等命名可能不同，必须与 checkpoint 训练配置一致。

\subsection{LongRoPE：逐维、分位置搜索缩放因子}

前述方法都用解析规则生成缩放。LongRoPE 进一步利用两种非均匀性\cite{zh:ding2024longrope}：
\begin{enumerate}
  \item 不同 RoPE 维度需要不同缩放因子 $\lambda_r$；
  \item 序列开头一段位置对插值更敏感，可设置保留区间 $n_{\mathrm{hat}}$。
\end{enumerate}
其一般化频率写作：
\begin{equation}
  \theta'_r=\frac{\theta_r}{\lambda_r},
  \label{zh:eq:longrope-scale}
\end{equation}
并用 evolutionary search 在约束空间中寻找 $\{\lambda_r\}$ 与 $n_{\mathrm{hat}}$。LongRoPE 先在 256K 长度适配，再基于该模型进行第二阶段搜索与插值，论文报告达到 2048K；同时为短上下文搜索另一套缩放配置以恢复原长度性能\cite{zh:ding2024longrope}。关键思想是：\term{扩展倍率不应被假设为跨维度、跨位置均匀。}

\subsection{LongRoPE2：从“理论周期”转向“有效训练周期”}

LongRoPE2 指出，低频/高维 RoPE 在预训练语料中不仅理论周期长，而且真正发生长距离依赖的样本稀少，导致其\term{有效相位范围}比理论预测更窄\cite{zh:longrope2}。因此，理论临界维度未必等于实际临界维度。它做了三项改进：
\begin{enumerate}
  \item 使用 needle-driven perplexity 指导 evolutionary search，联合寻找实际临界维度与逐维缩放因子；
  \item 对 OOD 维度施加单调非减的 $\lambda_r$ 约束；
  \item 混合上下文训练：短样本使用原始 RoPE，长样本使用 rescaled RoPE。
\end{enumerate}
混合训练目标可抽象为：
\begin{equation}
  \mathcal{L}=\eta\,\mathcal{L}_{\mathrm{short}}(\theta_{\mathrm{orig}})
  +(1-\eta)\,\mathcal{L}_{\mathrm{long}}(\theta_{\mathrm{scaled}}),
  \label{zh:eq:mixed-context-loss}
\end{equation}
它直接处理“长上下文适配”和“短上下文保持”之间的冲突。论文在 LLaMA3-8B 与 Phi3-mini 上报告 128K 有效上下文与较高短上下文保持率；这些数字是论文实验结论，不应外推为所有模型的默认保证\cite{zh:longrope2}。

\subsection{方法对比与适用条件}

\begin{table}[htbp]
\centering
\small
\begin{tabularx}{\linewidth}{@{}p{2.4cm}p{3.0cm}p{3.2cm}X@{}}
\toprule
方法 & 缩放对象 & 主要优点 & 主要代价/风险 \\
\midrule
PI & 所有位置/频率统一除以 $s$ & 稳定、实现最简单 & 高频局部分辨率下降，通常需微调 \\
NTK-aware & 修改 base，逐维非均匀缩放 & 保留最高频局部结构 & 社区经验起源，超参依模型 \\
Dynamic NTK & 倍率随当前长度变化 & 短输入保持原 RoPE & 与旋转后 KV Cache 冲突 \\
NTK-by-parts & 按旋转圈数分区插值 & 高频保留、低频防 OOD & 需选择 $\alpha,\beta$ \\
YaRN & NTK-by-parts + 温度校准 & 长短程折中更完整 & 必须匹配训练 recipe 与 attention factor \\
LongRoPE & 搜索逐维因子与位置保留区 & 能利用非均匀性，支持渐进扩长 & 搜索与验证成本更高 \\
LongRoPE2 & 搜索有效临界维度 + 混合训练 & 同时优化有效长度和短程保持 & 需要专门数据、搜索和 mid-training \\
\bottomrule
\end{tabularx}
\caption{主要 RoPE 扩展方法的统一比较。}
\label{zh:tab:rope-scaling}
\end{table}

\noteBox{\textbf{实践建议：}对已有 checkpoint，应复用其原生 RoPE 配置和训练 recipe；\textnormal{rope\_theta}、PI factor、YaRN factor 与 LongRoPE 的逐维因子并不等价。从头训练长上下文模型时，需要联合设计 RoPE scaling、长短样本 mix、position-id 语义和评估矩阵。}

\section{另一种设计：ALiBi 的距离线性偏置}

ALiBi 不旋转向量，而是在第 $h$ 个 head 的注意力分数中加入线性距离惩罚\cite{zh:press2022}：
\begin{equation}
  s_{ij}^{(h)}=
  \frac{q_i^{(h)\top}k_j^{(h)}}{\sqrt{d_h}}
  -m_h|i-j|.
  \label{zh:eq:alibi}
\end{equation}
不同 head 使用不同斜率 $m_h$。它形式简单、没有位置表，并显式注入 recency bias；但其归纳偏置与 RoPE 的多频率相位不同，是相对位置 bias 分支而非 RoPE 扩展。

\section{三类位置机制的统一视角}

三类方法分别修改输入、token 对分数和 Query-Key 几何关系：
\begin{align}
  x_i'&=x_i+p_i, &&\text{绝对位置}, \label{zh:eq:unified-absolute}\\
  s_{ij}&=\frac{q_i^\top k_j}{\sqrt{d_h}}+b(i-j), &&\text{相对位置 bias}, \label{zh:eq:unified-relative}\\
  s_{ij}&=\frac{(R_iq_i)^\top(R_jk_j)}{\sqrt{d_h}}
  =\frac{q_i^\top R_{j-i}k_j}{\sqrt{d_h}}, &&\text{RoPE}. \label{zh:eq:unified-rope}
\end{align}

\begin{table}[htbp]
\centering
\footnotesize
\begin{tabularx}{\linewidth}{@{}p{2.2cm}p{1.5cm}p{1.8cm}p{1.8cm}X@{}}
\toprule
机制 & 注入位置 & 相对距离 & 超长位置可计算 & 可靠外推条件 \\
\midrule
Learned Absolute & 输入 & 不显式 & 通常否 & 受位置表与训练长度限制 \\
Sinusoidal & 输入 & 几何中隐含 & 是 & 可计算不等于可利用 \\
Relative Bias & logits/K/V & 显式 & 依分桶而定 & 取决于分桶与训练 \\
RoPE & Q/K & 点积中显式 & 是 & 通常需 scaling 与适配训练 \\
ALiBi & logits & 显式 & 是 & 具有强 recency bias \\
\bottomrule
\end{tabularx}
\caption{位置机制的结构比较。}
\label{zh:tab:position-compare}
\end{table}

\section{工程实验与评估建议}

\subsection{配置必须随 checkpoint 固化}

对 decoder-only LLM，至少应保存并验证：
\begin{itemize}
  \item \textnormal{rope\_theta} 或 base frequency；
  \item rotary dimension：全维或 partial rotary；
  \item 维度配对布局；
  \item 原始训练长度与目标长度；
  \item scaling 类型、factor、attention factor 与逐维参数；
  \item packing、prefix cache、sliding window 下 position id 的定义。
\end{itemize}

\subsection{长上下文评估不能只看“能否输入”}

建议至少报告：
\begin{enumerate}
  \item 短上下文保持率：扩长前后原长度 PPL 与 benchmark；
  \item 按 token position、依赖跨度分桶的 PPL；
  \item needle/passkey 的长度--深度二维网格；
  \item RULER 类多任务长上下文评估，而非单一检索；
  \item 有效上下文：性能开始显著退化的位置，而非配置最大长度；
  \item prefill/decode latency、峰值显存、KV Cache 占用与吞吐。
\end{enumerate}

代码模型还应补充 definition-use 跨度、跨文件 import/call graph 定位、同名符号消歧、repo-level completion 与 patch correctness，并验证扩长后 HumanEval 类短程语法和算法能力是否退化。

\section{常见误解}

\begin{enumerate}
  \item \textbf{Self-Attention 完全看不见顺序。}更准确地说，不含位置机制的 Self-Attention 对输入置换等变。causal mask 提供可见方向，但不等于精细距离表示。
  \item \textbf{正弦编码是绝对位置，所以不含相对信息。}式~\eqref{zh:eq:sin-shift} 与式~\eqref{zh:eq:sin-dot} 表明固定偏移对应固定旋转，点积只依赖 $m-n$。
  \item \textbf{RoPE 是相对位置编码，因此不编码绝对位置。}单个 $R_mq_m$ 明确依赖 $m$；只有点积时绝对相位才相消。
  \item \textbf{位置函数能算到 128K，模型就有 128K 上下文。}配置长度、数值可运行长度与有效上下文长度是三件事。
  \item \textbf{RoPE 天然让注意力随距离单调下降。}式~\eqref{zh:eq:rope-oscillation} 是周期函数，不存在逐样本严格单调保证。
  \item \textbf{只调大 \textnormal{rope\_theta} 就完成扩长。}频率缩放只是第一步，还要验证模型适配、KV Cache 一致性、短程保持与有效上下文。
\end{enumerate}

\section{总结与展望}

绝对位置编码把坐标写入 token 表示，相对位置编码直接修改 token 对之间的关系，RoPE 则通过 Query-Key 旋转把绝对位置转化为点积中的相对相位。PI、NTK-aware、YaRN 和 LongRoPE 进一步处理频率缩放与长短上下文兼容问题；RoPE scaling law 则从周期覆盖出发，联系 base、训练长度、临界维度与外推范围。不过，现有位置机制仍有若干问题没有解决：

\begin{enumerate}
  \item \textbf{标称长度与有效长度不一致。}能接收 128K 或 1M token，不代表能稳定利用所有位置；有效上下文还受训练依赖跨度、优化过程和注意力稀释影响。
  \item \textbf{周期性带来相位混叠。}不同距离可能产生相近相位；多频率组合能够缓解，却不能从理论上消除这一问题。
  \item \textbf{扩长与短程能力仍有冲突。}插值会降低局部分辨率，逐维缩放和混合长度训练也缺少可跨模型迁移的稳定规律。
  \item \textbf{系统实现可能破坏位置语义。}packing、prefix cache、sliding window、speculative decoding 和 KV Cache 复用都可能造成隐蔽的 position-id 错位。
  \item \textbf{一维距离难以表达复杂结构。}代码的文件层级、调用图和 definition-use 关系，以及多模态数据的空间与时间结构，都超出单一序列坐标的表达范围。
  \item \textbf{位置编码与高效注意力缺少统一设计。}稀疏注意力、滑窗、KV 压缩和线性注意力会改变信息传播路径，位置机制需要与算子共同设计。
  \item \textbf{评估方法仍然偏弱。}needle-in-a-haystack 不能充分反映多跳推理、跨文件代码理解和长程因果依赖。
\end{enumerate}

后续研究不应只放大标称上下文，而应建立可预测、可训练、可缓存且能表达多维结构的位置机制，并用真实长依赖任务检验远距离信息利用率。对代码模型而言，definition-use、跨文件调用和 repository-level 修改比随机 needle 更接近实际需求。

\clearpage
\appendix
\section{代表性 LLM 的位置编码选择}

表~\ref{zh:tab:llm-position-encoding} 汇总公开论文、技术报告或官方开源配置中可以核验的位置机制，资料状态更新至 2026 年 8 月 7 日。表中的“上下文扩展”只记录位置编码相关做法，不代表模型仅凭该机制就能达到标称上下文长度。对于没有公开架构细节的模型，不能依据上下文长度或同系列旧模型反推出其位置编码。

\begin{longtable}{@{}p{2.7cm}p{3.6cm}p{8.0cm}@{}}
\caption{代表性语言模型公开披露的位置编码方式。}
\label{zh:tab:llm-position-encoding}\\
\toprule
模型/系列 & 位置机制 & 技术报告中的相关设计 \\
\midrule
\endfirsthead
\multicolumn{3}{l}{\small 表~\thetable（续）}\\
\toprule
模型/系列 & 位置机制 & 技术报告中的相关设计 \\
\midrule
\endhead
\midrule
\multicolumn{3}{r}{\small 下页续}\\
\endfoot
\bottomrule
\endlastfoot

GPT-2 / GPT-3 & 可学习绝对位置 embedding & GPT-2 使用可学习的位置 embedding；GPT-3 沿用 GPT-2 的基本 Transformer 架构，因此仍采用 learned absolute 方案\cite{zh:radford2019gpt2,zh:brown2020gpt3}。这类设计的最大长度通常受位置表和训练长度约束。\\

T5 & 分桶相对位置 bias & 每个 attention head 对相对距离桶学习标量偏置；近距离细分、远距离近似对数分桶\cite{zh:raffel2020}。\\

BLOOM & ALiBi & 不向 embedding 加位置向量，而是在注意力 logits 上施加按距离增长的线性负偏置\cite{zh:lescao2022bloom}。\\

Falcon & ALiBi & Falcon 系列技术报告采用 ALiBi；其消融实验同时比较了 ALiBi、URPE 与 RoPE\cite{zh:almazrouei2023falcon}。\\

PaLM & RoPE & 使用 RoPE 取代绝对或相对位置 embedding；报告将其长序列表现列为采用理由之一\cite{zh:chowdhery2022palm}。\\

LLaMA / Llama 2 & RoPE & LLaMA 以 RoPE 取代绝对位置 embedding；Llama 2 延续 RoPE，并把预训练上下文从 2K 扩为 4K\cite{zh:touvron2023llama,zh:touvron2023llama2}。\\

Code Llama & RoPE，增大基数 & 在 Llama 2 基础上把 RoPE base 调为 $10^6$，使用 16K 序列继续训练；报告研究了向 100K 的外推\cite{zh:roziere2023codellama}。\\

Llama 3 & RoPE，$\theta=500{,}000$ & 技术报告把 RoPE base 提高到 500K，用于改善长上下文支持；该配置对 8B、70B 与 405B 模型一致\cite{zh:dubey2024llama3}。\\

Qwen2 & RoPE + YaRN + DCA & 预训练后期把上下文从 4K 扩到 32K，并把 RoPE base 从 10K 调为 $10^6$；推理侧结合 YaRN 与 Dual Chunk Attention 支持更长输入\cite{zh:yang2024qwen2}。\\

Qwen3 / Qwen3.6 & RoPE 系列 + YaRN & Qwen3 的官方配置使用 $\theta=10^6$，可通过 YaRN 从原生 32K 扩展到 131K\cite{zh:yang2025qwen3}。Qwen3.6 进一步采用全注意力与线性注意力混合结构；其官方配置在全注意力层使用部分旋转的 M-RoPE，$\theta=10^7$，原生支持 262K，并给出使用 YaRN 扩展至约 1M 的配置\cite{zh:qwen36config}。\\

Gemma & RoPE & Gemma 在每层使用 rotary positional embeddings，而不是绝对位置 embedding；基础模型训练长度为 8K\cite{zh:gemmateam2024}。\\

Phi-3 / Phi-3.5 & RoPE 系架构 + LongRoPE & Phi-3-mini 采用与 Llama 2 相近的 block；128K 版本通过 LongRoPE 扩展，Phi-3.5 继续使用 LongRoPE 与混合上下文训练\cite{zh:abdin2024phi3}。\\

DeepSeek-V2 / V3 & MLA 中的 decoupled RoPE & 为兼容低秩 KV 压缩，将携带 RoPE 的 Query/Key 子空间与内容压缩子空间解耦；V3 延续该 MLA 设计\cite{zh:deepseek2024v2,zh:deepseek2024v3}。\\

LongCat 系列 & MLA 中的 decoupled RoPE & LongCat-Flash、LongCat-Flash-Thinking 与 2601 版本沿用 MLA，把 Query/Key 分成 RoPE 与 NoPE 子空间。初版 Flash 的官方配置为 64 维 RoPE 子空间、128 维 NoPE 子空间，$\theta=10^7$，最大位置为 131K；2601 的公开配置将 $\theta$ 设为 $10^6$\cite{zh:longcatflash,zh:longcatthinking2601,zh:longcatconfigs}。\\

Kimi K3 & NoPE + Kimi Delta Attention & 不向 MLA 的 Query/Key 注入显式位置编码；位置敏感性由 KDA 的递归门控与衰减机制隐式提供。技术报告指出该设计无需 RoPE rescaling 或 interpolation 即可继续训练到 1M 上下文\cite{zh:kimik3}。\\

GLM-5 & MLA/DSA 中的 decoupled RoPE & GLM-5 以 DSA 稀疏化 MLA 的注意力计算；官方配置保留 64 维 RoPE 子空间与 192 维 NoPE 子空间，并设置 $\theta=10^6$。DSA 改变稀疏检索路径，但没有取消 RoPE\cite{zh:glm5,zh:glm5config}。\\

Qwen3.8-Max & 未公开 & 官方服务文档已列出 Qwen3.8-Max，但截至本文资料截止日尚无公开技术报告、权重或配置可以确认位置机制。因此不能直接套用 Qwen3/Qwen3.6 的 RoPE 方案\cite{zh:qwen38docs}。\\

GPT-4 & 未公开 & GPT-4 技术报告未披露模型结构、位置编码或 RoPE 配置，因此无法从公开材料确定其位置机制\cite{zh:openai2023gpt4}。\\

\end{longtable}

从公开资料可以看到，decoder-only 开源模型在 2022 年以后明显集中到 RoPE，但并未完全收敛：BLOOM 与 Falcon 使用 ALiBi，T5 系模型采用相对位置 bias，DeepSeek、LongCat 与 GLM-5 为适配 MLA 或稀疏注意力而拆分 RoPE/NoPE 子空间，Kimi K3 则转向由 KDA 隐式提供位置信息的 NoPE 方案。所谓“使用 RoPE”也不是一个完整配置，base、rotary dimension、scaling、训练长度和 position-id 语义都会改变实际行为；对于 Qwen3.8-Max 这类未披露架构的模型，表中保留“未知”比沿用系列历史配置更可靠。

\end{document}